\def\year{2019}\relax
\documentclass[letterpaper]{article}
\usepackage{aaai_mod}
\usepackage{times}
\usepackage{helvet}
\usepackage{courier}
\usepackage{url}
\usepackage{graphicx}
\usepackage{algorithm}
\usepackage{algpseudocode}
\usepackage{listings}
\usepackage{refcount}
\usepackage[dash,dot]{dashundergaps}
\usepackage{stmaryrd}
\usepackage[usenames, dvipsnames]{xcolor}
\usepackage{latexsym}
\usepackage{bm}
\usepackage{amssymb}
\usepackage{amsmath}
\usepackage{multirow}
\usepackage[utf8]{inputenc}
\usepackage[titletoc,toc,title]{appendix}
\allowdisplaybreaks
\usepackage{tikz}
\usetikzlibrary{arrows.meta,matrix,decorations.pathreplacing}
\usepackage{dblfloatfix}

\title{Optimizing Segmentation Granularity for Neural Machine Translation}

\author{Elizabeth Salesky$^1$, Andrew Runge$^1$, Alex Coda$^1$, Jan Niehues$^2$, Graham Neubig$^1$ \\
  $^1$Carnegie Mellon University, Pittsburgh, PA, USA\\
  {\tt  \{esalesky,ajrunge,acoda,gneubig\}@cmu.edu} \\
  $^2$Karlsruhe Institute of Technology, Karlsruhe, Germany \\
 {\tt jan.niehues@kit.edu} \\ }

\begin{document}
\maketitle

\begin{abstract}
In neural machine translation (NMT), it is has become standard to translate using subword units to allow for an open vocabulary and improve accuracy on infrequent words.
Byte-pair encoding (BPE) and its variants are the predominant approach to generating these subwords, as they are unsupervised, resource-free, and empirically effective.
However, the \emph{granularity} of these subword units is a hyperparameter to be tuned for each language and task, using methods such as grid search.
Tuning may be done inexhaustively or skipped entirely due to resource constraints, leading to sub-optimal performance.
In this paper, we propose a method to automatically tune this parameter using only one training pass.
We incrementally introduce new vocabulary online based on the held-out validation loss, beginning with smaller, general subwords and adding larger, more specific units over the course of training.
Our method matches the results found with grid search, optimizing segmentation granularity without any additional training time.
We also show benefits in training efficiency and performance improvements for rare words due to the way embeddings for larger units are incrementally constructed by combining those from smaller units.%
\end{abstract}

\section{Introduction}

A key issue NMT systems face in training is vocabulary sparsity; many words are seen only a few times and in insufficient contexts for neural models to learn to handle them properly. This problem is compounded in morphologically rich languages like Czech, German, or Turkish, where a greater number of word forms appear very few times or not at all in the available training data.
Translating into these languages is particularly challenging, both due to difficulties in accurately generating rare word forms and memory restrictions due to a higher number of parameters for the output vocabulary.
As a result, it is typically necessary to limit target vocabulary size. 
To address this, a large body of past work has explored techniques to segment words into smaller units, including morphological analysis, character-based models, and subword models. 

The current standard approach to handle large, sparse vocabularies is byte-pair encoding (BPE), which was first applied to NMT by Sennrich et al. \shortcite{Sennrich2016a}. 
BPE is simple, unsupervised, does not require additional resources, and has been shown to improve performance.
BPE iteratively merges pairs of frequent character sequences to create subword units.
Translating with subwords allows less common or unseen word forms to be composed of multiple more common subwords, enabling models to translate to an open vocabulary.
However, the \emph{granularity} of the segmentation produced by BPE, determined by the number of merge operations, is a hyperparameter to be tuned for the language and corpus.
To do so requires separate full training runs, which is time and resource intensive, taking GPU-days or weeks.
Instead of performing this sweep, one setting is often held constant across many experiments and language pairs, or only few are compared, limiting the achievable BLEU score.

In this paper, we propose a method to tune segmentation granularity automatically as part of the training process. 
We do so by incrementally expanding the target vocabulary online to include longer, more specific subwords at intervals determined by loss on the held-out validation set.
This allows us to optimize the number of BPE merges without requiring additional training runs to sweep this parameter.
In order to ensure that vocabulary added online can be utilized effectively, we propose two methods to initialize new embeddings, and compare to random initialization as a control.
To isolate the effects of language and data size, we compare two datasets of different sizes for two language pairs.
We show that our method is able to match, in a single pass, the best performance achieved by a more expensive grid search over segmentation granularities. 
In addition, we show that our online training scheme improves training time and rare word prediction accuracy.

\section{Byte-Pair Encoding}
\label{bpe-def}

Segmentation into subwords is done as a pre-processing step for MT. 
For our segmentation strategy, we use byte-pair encoding (BPE) \cite{gage1994new}. 
BPE is a data compression algorithm that iteratively replaces pairs of bytes that frequently occur adjacently with an unused byte. 
This has been adapted for unsupervised word segmentation by Sennrich et al. \shortcite{Sennrich2016a}, where instead frequent pairs of adjacent characters and character sequences are replaced with a single longer sequence. 
Merges occur until a pre-specified number of merge operations, and then the resulting set of subwords is applied to the training corpus to create the final vocabulary for MT.
An example of how different numbers of merges affect a sentence from our Czech data can be seen below in Figure \ref{bpe czech sample}.
(`@@' is used to mark boundaries within words. This ensures that although translation is conducted using subwords, we can recover valid word forms for the final output after translation).
As shown in this example, BPE takes pairs of adjacent subwords and merges them into a single token. 
Merge operations build on each other: we see the subword produced in the merge on line 2 (\textit{Pohy@@}) is used in the subsequent merge on line 3, producing \textit{Pohybují}. 

\begin{figure}[!ht]
\centering
\begin{tabular}{cl}
\bf \# Operations & \bf Resulting Sentence \\
10k & \dashuline{Po@@~ hy@@~ bují} se nahoru a dolů . \\
30k & \dashuline{Pohy@@~ bují} se nahoru a dolů . \\
50k & \dashuline{Pohybují} se nahoru a dolů . \\
\end{tabular}
\caption{The Czech word \textit{`Pohybují'} is represented by different numbers of  subwords based on the number of BPE merge operations.}
\label{bpe czech sample}
\end{figure}

Using more merge operations reduces the number of subwords needed to represent each word, subsequently reducing the number of subwords the model must generate correctly in a row in order to correctly translate a word.
However, it also reduces the number of times each vocabulary item is present in the training corpus. 
Depending on its frequency, this can impact the model's ability to learn to translate it correctly. 
The choice of the number of merge operations is a balance between these two forces (frequency and length).

BPE merge operations are determined by a given corpus' statistics. 
Which character sequences are merged depends on the frequency of character sequences in the training data used to produce the subword vocabulary.
The optimal number of BPE merges and so the resulting vocabulary is therefore dependent on both language and dataset.
The number of merge operations to be used is determined experimentally by running full MT experiments using different subword vocabularies and comparing the resulting BLEU scores. 
This can be very costly, particularly using larger datasets where a single MT experiment can take days. 
It is common to use a relatively high value like 50k in place of tuning, leading to reasonable but non-optimal performance. 
Our method tunes the merge operations parameter, and therefore the vocabulary size and granularity, to achieve optimal performance without requiring these additional experiments.

\section{Incremental BPE}

We propose a method to optimize the number of BPE operations during training.
This enables optimization of this hyperparameter to maximize BLEU, without requiring additional resources or training time. 
To do so, we start training with a low number of BPE operations, and incrementally add vocabulary from additional BPE operations online. 
This method is simple but effective, making it easy to incorporate into a typical MT training pipeline.
Conducting this procedure effectively is determined by a \emph{merging schedule} and an \emph{initialization strategy} for new embeddings.
We will describe each of these in detail in turn.

\subsection{Merging Schedule}

For our experiments, we begin training with a small BPE subword inventory to be above character-level, but safely below the optimal segmentation level so that the vocabulary size will strictly increase to the optimal granularity.
When training plateaus with the current vocabulary as determined by the loss on a held-out validation set either increasing or decreasing within a small threshold, we increase the size of the target vocabulary to the next interval. 
Larger BPE inventories contain the subwords of smaller inventories for a fixed dataset, so this is a strictly increasing change.
The specifics of our merge procedure are described in Algorithm \ref{merge-algorithm}. 

\begin{algorithm}
\begin{algorithmic}
\State $\ell_{thresh} = 0.05$
\State $inc_{curr} = 0$
\State $incs$ = \{10k,20k,30k,40k,50k,60k\}
\State $burn\_in$ = 3 epochs \\
\Repeat~ After each training epoch $e$:
\State $\Delta_{\ell} =  |\ell_{e} - \ell_{e-1}|$
\If {$\ell_{e}>\ell_{e-1}$ or $\Delta_{\ell}\leq \ell_{thresh}$}
    \If {\text{not in $burn\_in$}}
    	\State $inc_{curr}$++
        \State $init\_inc(incs[inc_{curr}])$
    \EndIf
\EndIf
\Until convergence
\end{algorithmic}
\caption{Incremental BPE Schedule}
\label{merge-algorithm}
\end{algorithm}

We start with 10k BPE operations, and when the validation loss increases or decreases within our threshold, we add additional vocabulary in increments of 10k additional subwords ($incs$). 
We set the validation loss threshold ($\ell_{thresh}$) to 0.05 because this is the range within which we observe loss begin to converge. 
We use a burn-in period ($burn\_in$) after adding new vocabulary before allowing another increment to ensure the network has time to learn to use the new vocabulary effectively.
We observe that 3 burn-in epochs allow the model to stabilize the length of generated sentences, and so use this number in all of our experiments. 
This incremental process is integrated into the overall model training; we continue training and potentially incrementing until a maximum patience of 10 epochs from the first loss increase is hit.

\newcommand\tikzmark[1]{\tikz[remember picture] \node (#1) {};}
\begin{table*}[hb!]
\centering
\begin{tabular}{c|c|c|c|c|c|c|c|c|}
\cline{2-9}
\multirow{2}{*}{} & \multicolumn{2}{c|}{IWSLT'16} & \multicolumn{2}{c|}{WMT'15} & \multicolumn{2}{c|}{IWSLT'16} & \multicolumn{2}{c|}{WMT'14} \\ \cline{2-9} 
 & English & Czech & English & Czech & English & German & English & German\\ \hline
\multicolumn{1}{|c|}{Words} & 40,601 & 108,027 & 50,000* & 50,000* & 58,466 & 124,941 & 50,000* & 50,000* \\ \hline
\multicolumn{1}{|c|}{50k} & 33,225 & 48,628 & 49,173 & 49,850 & 44,375 & 48,548 & 51,246 & 51,412 \\ \cline{1-1} \cline{3-3} \cline{5-5} \cline{7-7} \cline{9-9}
\multicolumn{1}{|c|}{40k} & \tikzmark{a} & 39,582 & \tikzmark{c} & 40,903 & \tikzmark{e} & 39.257 & \tikzmark{g} & 41,658 \\ \cline{1-1} \cline{3-3} \cline{5-5} \cline{7-7} \cline{9-9}
\multicolumn{1}{|c|}{30k} & & 29,912 & & 30,991 & & 29,737 & & 31,824 \\ \cline{1-1} \cline{3-3} \cline{5-5} \cline{7-7} \cline{9-9}
\multicolumn{1}{|c|}{20k} & & 20,101 & & 21,043 & & 20,022 & & 21,910\\ \cline{1-1} \cline{3-3} \cline{5-5} \cline{7-7} \cline{9-9}
\multicolumn{1}{|c|}{10k} & \tikzmark{b} & 10,215 & \tikzmark{d} & 11,063 & \tikzmark{f} & 10,144 & \tikzmark{h} & 11,904 \\ \hline
\end{tabular}
\tikz[remember picture,overlay] \draw[->] (a.center -| b.center) -- (b.center);
\tikz[remember picture,overlay] \draw[->] (c.center -| d.center) -- (d.center);
\tikz[remember picture,overlay] \draw[->] (e.center -| f.center) -- (f.center);
\tikz[remember picture,overlay] \draw[->] (g.center -| h.center) -- (h.center);
\caption{Vocabulary sizes by number of BPE operations. We hold English vocabs constant when tuning BPE, using 50k BPE operations. For word-based WMT experiments we use the 50k most frequent words. }
\label{vocab_size}
\end{table*}

\subsection{Initialization Strategies}

When new vocabulary is introduced online, certain modifications to the network are required.
New entries must be added to the target embedding matrix, and correspondingly we extend the weight and bias tensors in the decoder output layer so the model is able to generate the new embeddings.
Proper initialization is important to allow the network to generate the new words with minimal additional training.
BPE operations take two adjacent subwords and merge them into one new subword. 
Motivated by this, when we initialize parameters (e.g. an embedding) for a new vocabulary item, we combine the trained parameters for its two component subwords.
We compare two initialization strategies to random initialization (\textbf{Rand}) as a control to demonstrate that their effectiveness: a naive average (\textbf{Avg}), and using an autoencoder (\textbf{AE}).

\subsubsection{Avg}
\lstset{basicstyle=\scriptsize\selectfont\ttfamily}

Our first strategy initializes the embeddings for each new vocabulary item by averaging the trained embeddings of their two component subwords, shown here in PyTorch:
\begin{lstlisting}
   # new_embed created from the two at indices x & y
   new_embed = torch.div(torch.add(embed.weight[x], 
                embed.weight[y]), 2)
   torch.cat((embed.weight[:].data, new_embed.data), 0)
\end{lstlisting}

\subsubsection{AE}

\begin{figure}[ht!]
  \centering
  \includegraphics[width=0.9\linewidth]{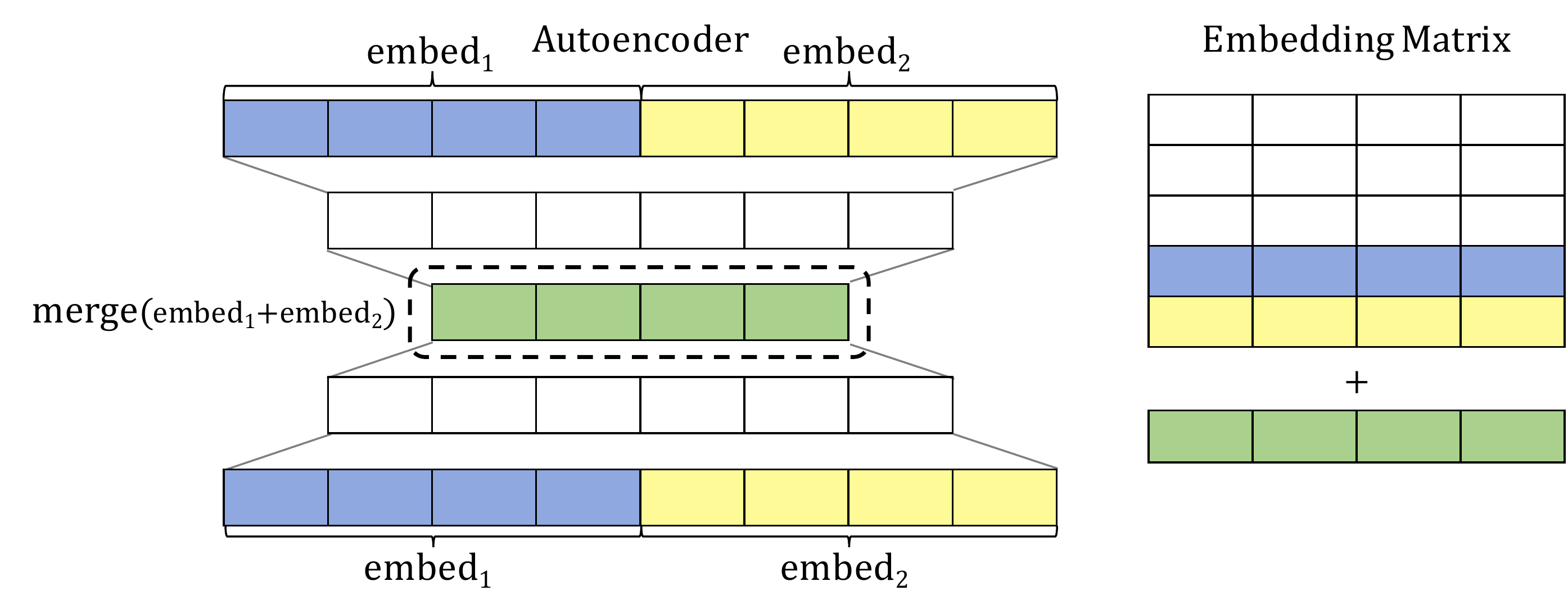}
  \caption{Autoencoder architecture for initialization.}
  \label{ae}
\end{figure}

Our second strategy uses an autoencoder to initialize new embeddings by compressing the trained embeddings of their component subwords to the size of a single embedding.
To do so, we concatenate the two component embeddings and use this as the source and target for an autoencoder.
We use three hidden layers with ReLU functions, and set the middle hidden layer to the dimension of an embedding. 
We use this compressed representation as the initialization for the new embedding (shown in Figure \ref{ae}).
Whenever new vocabulary is added, we pause the sequence-to-sequence model training, and continue the autoencoder training for 50 epochs.
We optimize using cross-entropy loss.
It has significantly fewer parameters, so training takes less than $30s$, making the change in training time to a full run insignificant.

\section{Experiments}

To evaluate our incremental BPE approach, we perform experiments on two dataset sizes for two different language pairs. 
We compare our method against grid searches over the number of BPE operations.
We use data from the IWSLT and WMT evaluation campaigns to be comparable to other work, and specifically the Czech and German tasks as morphologically-rich languages with a body of previous work.

\begin{table*}[ht!]
\centering
\begin{tabular}{c|c|c|c|c||c|c|c|c|}
\cline{2-9}
\multirow{2}{*}{} & \multicolumn{4}{c||}{\bf English-Czech} & \multicolumn{4}{c|}{\bf English-German} \\ \cline{2-9} 
\multirow{2}{*}{} & \multicolumn{2}{c|}{IWSLT'16} & \multicolumn{2}{c||}{WMT'15} & \multicolumn{2}{c|}{IWSLT'16} & \multicolumn{2}{c|}{WMT'14} \\ \cline{2-9} 
 & \texttt{dev} & \texttt{test} & \texttt{dev} & \texttt{test} & \texttt{dev} & \texttt{test} & \texttt{dev} & \texttt{test} \\ \hline
\multicolumn{1}{|c|}{10k}   & 13.35 & 13.51 & 9.75 & 9.60   & 21.13 & 22.07 & 13.34 & 12.10 \\ \hline
\multicolumn{1}{|c|}{20k}   & \bf 13.48 & 13.60 & 9.92 & 9.61   & 21.19 & 22.45 & 13.71 & 12.61 \\ \hline
\multicolumn{1}{|c|}{30k}   & 13.20 & 13.35 & 10.22 & 10.00 & 21.28 & 22.17 & 13.73 & 12.50\\ \hline
\multicolumn{1}{|c|}{40k}   & 13.21 & 13.60 & \bf 10.37 & \bf 10.32 & \bf 21.46 & \bf 22.51 & \bf 13.84 & \bf 12.83 \\ \hline
\multicolumn{1}{|c|}{50k}   & 13.09 & 13.56 & 10.30 & 10.17 & 21.31 & 21.89 & 13.78 & 12.71 \\ \hline
\multicolumn{1}{|c|}{60k}   & ----- & ----- & ----- & ----- & ----- & ----- & 13.70 & 12.48 \\ \hline \hline
\multicolumn{1}{|c|}{Words} & 13.37 & 13.48 & 9.84 & 9.59 & 21.31 & 22.12 & 12.48 & 11.06 \\ \hline
\end{tabular}
\caption{Results from BPE sweeps in BLEU. Results from the smaller IWSLT'16 datasets are averages over three runs.}
\label{all_bpe_sweep}
\end{table*}
\begin{table*}[hb!]
\centering
\begin{tabular}{c|c|c|c|c|c|c||c|c|c|c|c|c|}
\cline{2-13}
\multirow{2}{*}{} & \multicolumn{6}{c||}{\bf English-Czech} & \multicolumn{6}{c|}{\bf English-German} \\ \cline{2-13} 
\multirow{2}{*}{} & \multicolumn{3}{c|}{IWSLT'16} & \multicolumn{3}{c||}{WMT'15} & \multicolumn{3}{c|}{IWSLT'16} & \multicolumn{3}{c|}{WMT'14} \\ \cline{2-13} 
 & \texttt{dev} & \texttt{test} & \texttt{BPE} & \texttt{dev} & \texttt{test} & \texttt{BPE} & \texttt{dev} & \texttt{test} & \texttt{BPE} & \texttt{dev} & \texttt{test} & \texttt{BPE} \\ \hline
\multicolumn{1}{|c|}{\bf Rand} & 13.58 & 13.13 & 30k & 10.16 & 10.06 & 30k & 21.22 & 22.28 & 30k & 13.74 & 12.76 & 20k  \\ \hline\hline
\multicolumn{1}{|c|}{\bf Avg}  & 13.38 & 13.82 & 30k & 10.33 & 10.01 & 20k & 21.11 & 22.04 & 30k & 13.81 & 12.54 & 30k \\ \hline 
\multicolumn{1}{|c|}{\bf AE}   & 13.74 & 13.69 & 20k & 10.10 & 10.16 & 30k & 21.62 & 22.53 & 30k & 13.82 & 12.77 & 30k \\ \hline\hline
\multicolumn{1}{|c|}{\bf Sweep}& 13.48 & 13.60 & 20k & 10.37 & 10.32 & 40k & 21.46 & 22.51 & 40k & 13.84 & 12.83 & 40k \\ \hline
\end{tabular}
\caption{Incremental BPE Results in BLEU, compared to the best sweep result. \texttt{BPE} is the BPE inventory of the final model.}
\label{all_incremental}
\end{table*}

\subsection{Data}
We translate from English into two morphologically-rich languages, Czech and German.
To isolate the effects of language and dataset size, we compare four conditions.
For our initial experiments, we use the two smaller IWSLT'16 English-Czech and English-German datasets, with the \texttt{tst2012} and \texttt{tst2013} for validation (\texttt{dev}) and \texttt{test}.
These have 105k and 185k sentences of training data, respectively. 
We then compare on a larger dataset, using 1M sentence subsets from the Stanford preprocessed WMT data for both language pairs.\footnote{https://nlp.stanford.edu/projects/nmt/} 
For these experiments, we use \texttt{newstest2013} and \texttt{newstest2014} for \texttt{dev} and \texttt{test}. 
We tokenize with the Moses tokenizer.\footnote{\label{moses}https://github.com/moses-smt/mosesdecoder/}
All BLEU scores from IWSLT experiments are averages of three runs to account for optimizer instability on the smaller datasets.
Using WMT data, all scores are from single experiments.
All BLEU scores are generated using tokenized lowercased \texttt{multi-bleu}\footnotemark[\getrefnumber{moses}]. 
We hold the English vocabulary constant across experiments when tuning target-side BPE to isolate the effects of changing target-side segmentation; for all segmentation experiments, the English side uses 50k BPE operations.
Table \ref{vocab_size} shows the vocabulary sizes for these datasets by number of BPE operations.

\subsection{Model}
For our MT model, we use a basic seq2seq attentional model \cite{Bahdanau2014} implemented in PyTorch \cite{paszke2017automatic}.
We use an encoder with a single bidirectional GRU hidden layer. 
For the decoder, we implemented a single-layer conditional GRU with MLP attention and the deep output layer of Sennrich et al. \shortcite{Sennrich2017}. 
The embedding layers are 500 dimensional, and all hidden layers are size 1024. 
We create batches of size 60 with source sentences of the same length and target sentences of as similar length as possible.
We shuffle batch order every epoch.
We decode with beam width 5, normalizing final sentence scores by the average sentence length ratio $\gamma$ between source and target to force generated length to be close to expected length.
\begin{gather*}
\begin{split}
Score_{final} = Score * \Big( 1 + \frac{abs(\gamma * |src| - |gen|)}{|gen|} \Big)
\end{split}
\end{gather*}

\subsection{BPE Sweep Results}
\label{sweep_results}

First, to emphasize the importance of BPE granularity on translation accuracy, we perform a sweep over various numbers of BPE merge operations.
The results of the BPE sweep are shown in Table \ref{all_bpe_sweep}.

We see that performance trends slightly vary according to both language and dataset size. 
On the larger WMT datasets, increasing vocabulary size has a positive impact on BLEU score, until a point. 
System performance peaks at 40k and then begins to decrease for both WMT English-Czech and English-German.
There is also a significant difference between BPE operations for both language pairs using more training data; selecting a non-optimal setting has a large impact, up to 0.6 BLEU.
Comparing the two datasets for a single language pair, we see that the best setting for one condition is not necessarily the best for another. 
While both English-German datasets see their best results with 40k BPE, for Czech, the best setting for IWSLT'16 \texttt{dev} is 20k, while for WMT'15, 20k is 0.45 worse than 40k.
These results imply optimal performance on smaller datasets may occur with fewer BPE units.
BPE also provides a clear improvement over words, which are non-tenable to run without limiting the vocabulary for datasets of this scale. BPE allows an open vocabulary, and so removes the need to replace words with unknown tokens.

On the smaller IWSLT datasets, the correlation between BPE and performance is less stable. 
However, we identify several important trends.
For Czech, the best BPE granularity on \texttt{dev} is lower than the best on the larger WMT data.
We verify that the best result (20k) is statistically significant over the larger inventories (30k+) using bootstrap resampling,\footnote{https://github.com/jhclark/multeval} $p<0.05$ \cite{clark2011bootstrap}.
The best results are also significant over words; while it is possible to use word-based vocabularies on smaller datasets, the representation is sparser. Segmentation allows each type to be seen more times, important with less training data.
We see higher standard deviation on these smaller datasets, which makes this parameter more difficult to tune. 
On \texttt{test}, though the average over three runs is the same (13.60), the best 20k run is statistically different from the best 40k run and better by 0.17 BLEU.
For IWSLT German, best vocabulary is the same as WMT. 
However, there is a significant drop in performance (0.62) on \texttt{test} between 40k and 50k.
These results emphasize the importance and difficulty of tuning this parameter; here, using 50k, a common BPE setting, would have a large performance cost.

As evidenced by these experiments, choosing the correct segmentation granularity can affect translation accuracy, and this parameter is hard to tune well because it can vary by language and dataset.
Further, running these sweeps is costly, particularly when taking into account deviation between initializations on smaller datasets.
In the following experiments, we examine the ability of our method to automatically tune this granularity without costly parameter sweeps.


\subsection{Incremental BPE Results}
\label{inc_bpe_results}

The results of our experiments using incremental BPE are shown in Table \ref{all_incremental}.
The \texttt{BPE} column shows the size of the BPE inventory used by the best model at the end of training.
As shown in this table, our \textbf{AE} models match the BLEU of the best model from the BPE sweep, with the exception of WMT Czech, discussed below.
The \textbf{AE} typically performs better than both \textbf{Avg} and \textbf{Rand}. 
Further, the \textbf{AE} initialization method converges faster than \textbf{Avg} and \textbf{Rand}, explored in the next section.

On WMT Czech, the incremental results approach but do not quite meet the best results from the sweep.
For this dataset, the best sweep result occurs with 40k BPE units, while the best incremental  results use only 20k or 30k.
With this data, the best \texttt{dev} BLEU was reached as early as epoch 8, not allowing the incremental models sufficient time to add larger (30k+) BPE inventories before overtraining.
It may be that a lower number of burn-in epochs or a higher \texttt{dev} loss threshold would allow models to add this vocabulary earlier.
If these were added earlier in training, perhaps results would continue to improve.

The best incremental model results typically use a smaller segmentation granularity than the best sweep results. 
However, each has the same final vocabulary as an experiment from the sweep; for all four datasets, we see that \textbf{AE} does the same or better than the sweep experiment with the same vocabulary. 
We verify using bootstrap resampling that the \textbf{AE} outputs are statistically significant over the sweep results using the same vocabulary, $p<0.05$.
This is inconsistently true for \textbf{Avg} and \textbf{Rand}, particularly on the smaller IWSLT datasets.
Further, on German, the difference between \textbf{Avg} and \textbf{Rand} initialization shrinks.
This indicates that the more naive \textbf{Avg} initialization strategy is not always more effective than random.
We investigate the \textbf{AE} results further with respect to convergence and rare words below.

We compare our incremental models across two language pairs and dataset sizes. 
We find that the BPE setting that is most optimal for one condition may not be the best for another; however, our best incremental models are able to yield these results without requiring additional experiments to tune this setting.
Performing a sweep with a single training run per setting on even our smallest dataset took 34.5 GPU-hours.
By comparison, our incremental model required just a single training run, 7.1 hours, providing significant time savings over the manual tuning process while matching the performance of any single best BPE inventory. 
This method is simple and easy to implement, and enables us to get the most out of BPE segmentation without the overhead of tuning. 

Below in the training analysis we explore the training process with our model further to better understand these results.
In the further analysis section we look into additional benefits of our training method.

\section{Incremental BPE Analysis}
\label{analysis}

We now use our IWSLT English-Czech models to evaluate additional impact from our incremental training method.

\subsection{Impact on Training and Convergence}
\label{training_analysis}

Here we explore how making incremental additions to the model's vocabulary online affects the training process to better understand our results.

\begin{figure}[ht!]
  \centering
  \includegraphics[width=1.0\linewidth]{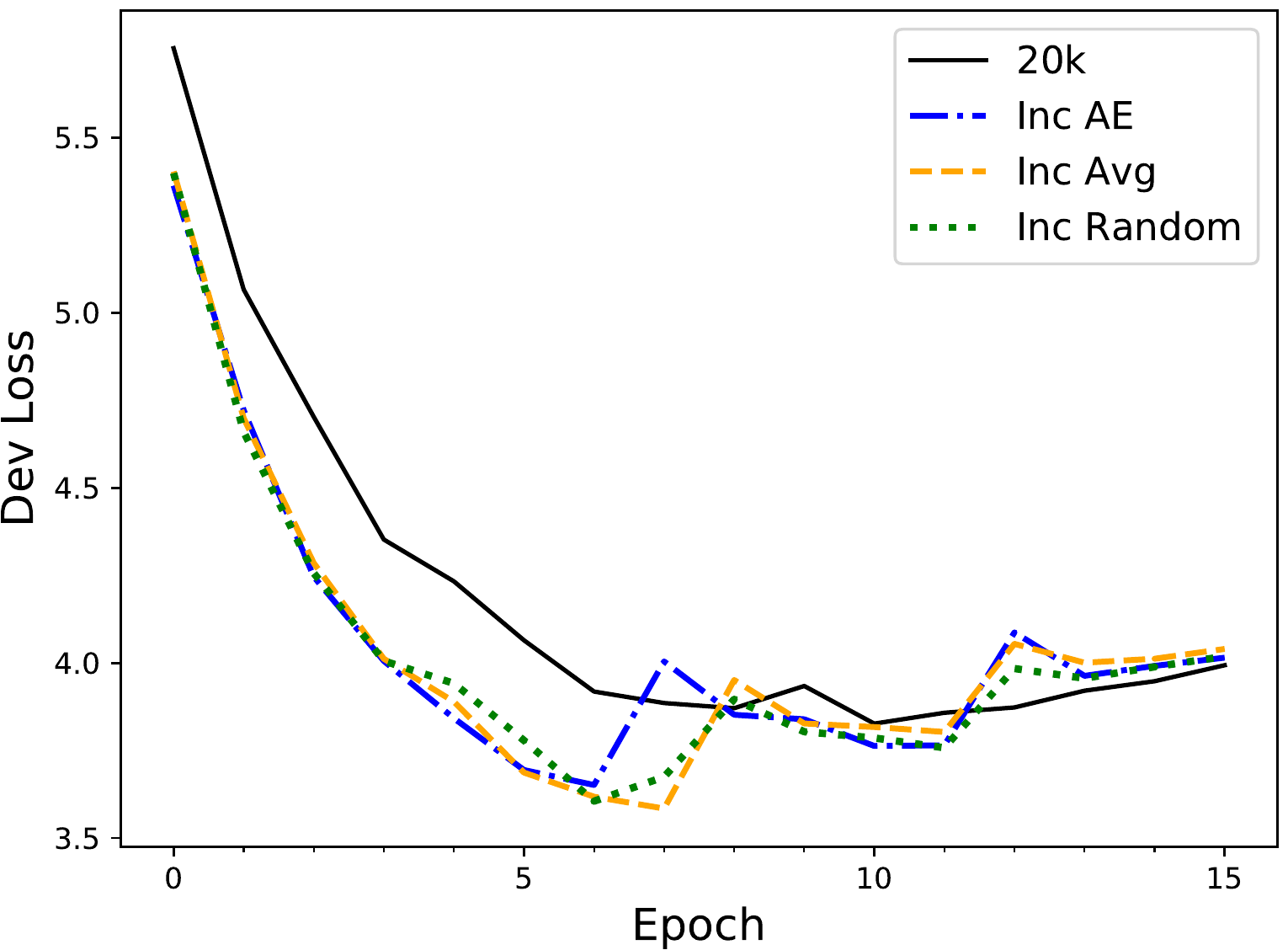}
  \caption{Dev loss comparison between best sweep experiment (20k) and incremental BPE models on IWSLT'16 English-Czech}
  \label{dev_ae_loss}
\end{figure}

First, we look at our loss when introducing new vocabulary with the incremental BPE models. 
We see the effects of adding new vocab on \texttt{dev} loss in Figure \ref{dev_ae_loss}, which compares the incremental BPE systems to the best result from the IWSLT'16 English-Czech BPE sweep.
Loss spikes at each incremental vocabulary introduction, but then continues to decrease, suggesting that the model learns to use the newly introduced vocabulary as it continues training. 
The initial gap in Figure \ref{dev_ae_loss} is because losses are not directly comparable between different vocabulary sizes, and the incremental models begin training with 10k vocabulary as opposed to 20k. 
After adding new vocabulary is therefore expected that the curves do not drop to their previous lowest loss.
Their final losses are similar to the best constant vocabulary experiment with the same vocabulary (Table \ref{all_bpe_sweep}).
As discussed with the incremental results, this figure also makes it clearer that \textbf{Avg} is more similar to \textbf{Rand} than \textbf{AE}.

We also notice that the incremental BPE experiments optimize slightly faster.
This may be because different segmentation granularities optimize at different rates. 
Comparing using the sweep experiments, we see that different BPE inventories peak at slightly different points in training. 
Looking at performance over training, we also find that smaller vocabularies (e.g. 10k, 20k) are less stable from epoch to epoch initially (15\% greater standard deviation), but can reach higher BLEU than larger vocabs later in training as shown in Table \ref{all_bpe_sweep}.

\begin{figure}[ht!]
  \centering
  \includegraphics[width=1.0\linewidth]{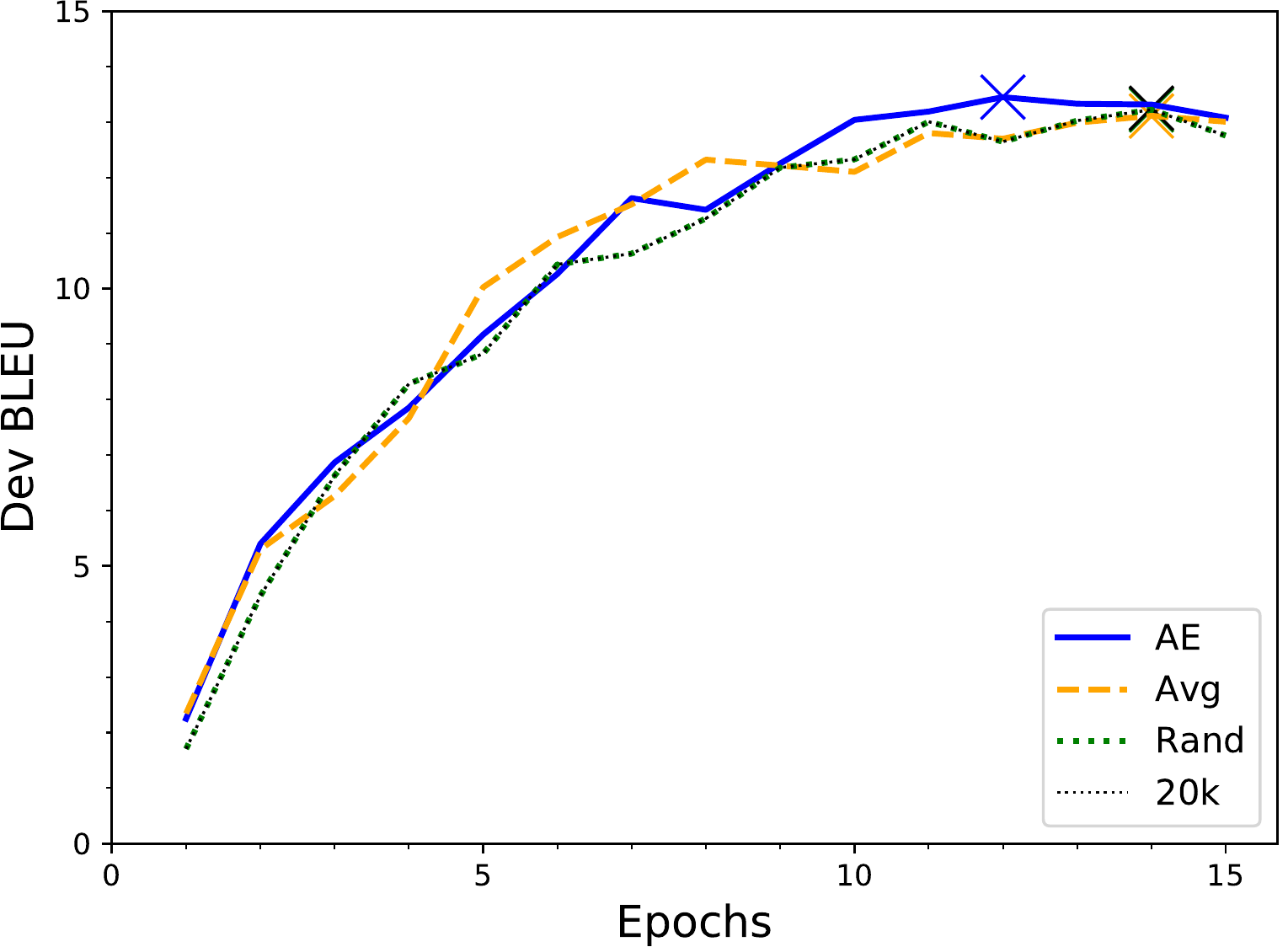}
  \caption{Dev BLEU over training using incremental BPE, IWSLT'16 English-Czech.}
  \label{avg_ae_plot}
\end{figure}

For our incremental experiments, Figure \ref{avg_ae_plot} shows the same comparison across our incremental BPE models, with the sweep 20k as a direct comparison.
Comparing the different incremental BPE initialization methods, the \textbf{AE} embedding initialization performs better than the \textbf{Avg} initialization, which is slightly worse than \textbf{Rand}.
We see that \textbf{Avg} and \textbf{Rand} begin to plateau earlier in training but reach their best epochs later than \textbf{AE}, while \textbf{AE} reaches higher overall BLEU scores.
This coincides with the overall trend we see across all four dataset conditions: the \textbf{AE} incremental BPE systems reach their best \texttt{dev} performance on average 2 epochs earlier than our other models (marked in Figure \ref{avg_ae_plot} by $`\times'$).
\textbf{AE} also uses a smaller final vocabulary and requires fewer epochs to converge than \textbf{Avg} or random initialization, suggesting it is a more effective way to initialize the new embeddings.
We also see that the 20k experiment is on average almost indistinguishable from the incremental \textbf{Rand} model's training trajectory, suggesting incremental training with strategic initialization benefits training and convergence.

Faster convergence may be because incremental training allows the system to initially train with more general vocabulary, each seen more frequently and so updated more times. 
When new vocabulary is introduced it is initialized from these embeddings, effectively `pretraining' them. 
Rarer embeddings are essentially updated more times than they would be in a typical training procedure.
The effects are felt in convergence time and performance on rare words, analyzed below.

\subsection{Rare Word F1}

\begin{figure}[ht!]
  \centering
  \includegraphics[width=1.0\linewidth]{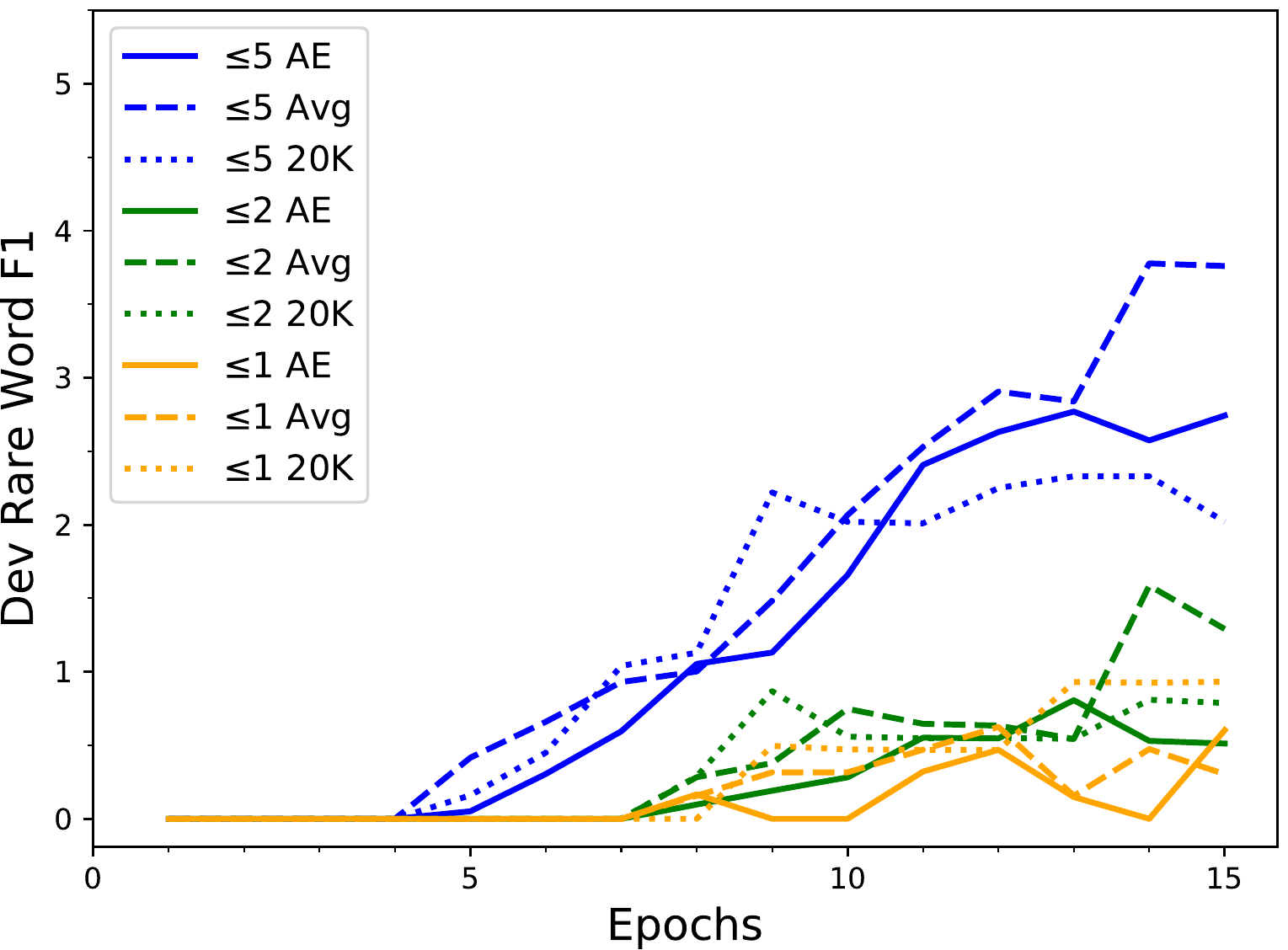}
  \caption{F1 of Rare Words Appearing $1,2,5\times$}
  \label{rare}
\end{figure}

To evaluate the impact of our incremental models on less frequent vocabulary, we look at rare words appearing \{1,2,5\}$\times$ in training, shown in Figure \ref{rare}.
We measure this through unigram F1, which we calculate as the harmonic mean of unigram precision and recall, as in \cite{Sennrich2016a}.
We see that the incremental BPE models are similar to each other and improve on rare word F1 over the constant 20k model for tokens appearing $5\times$.
This improvement is likely due to our initialization strategies, which provide some `pretraining.'
For rare vocabulary introduced incrementally, instead of having few training examples, they benefit from their component subwords' training examples as well.
For example, \textit{roda} occurs only $4\times$ but was introduced incrementally in the 20k inventory. It was created by merging \textit{ro} and \textit{da}, which occur 1718 and 2016 times, respectively. Because our initialization scheme makes use of the trained embeddings for \textit{ro} and \textit{da} to initialize \textit{roda}, it receives the benefit of much higher data frequency.
For vocabulary only appearing 1 or 2 times, results are inconsistent; this vocabulary may still not appear enough times for our method to provide additional benefit.
The composition of rare words changes with incremental BPE; they are composed of subword units from different BPE vocabularies.
Rare words are generated using on average ${\sim}12\%$ fewer subwords by epoch 15 with the incremental systems than the 20k sweep experiment.
This may make this vocabulary easier to generate correctly.

\subsection{Usage of Vocabulary Introduced Online}

Our results suggest the incremental BPE systems learn to generate the vocabulary introduced online. 
Looking at an example from the Czech data in Figure \ref{australii example}, we see the word \textit{Austrálii} has been split into three subwords with 5k operations, two of which are merged into the new vocabulary term \textit{Austrá@@} with 10k operations. 
An incremental BPE system starting from 5k and later adding the 10k vocab is correctly able to generate this newly introduced term. 

\begin{figure}[ht]
\centering
\begin{tabular}{cl}
\bf Dataset & \bf Text \\ \hline
\it Reference & útěk její rodiny a nový život v \dashuline{Austrálii} \\[4pt]
5k Output & za@@ chra@@ ň@@ ovali své rodiny \\
   & a nové život v \dashuline{Au@@ strá@@ lii} \\[4pt]
10k Output & vy@@ dání její rodiny a nová život \\
    & v \dashuline{Austrá@@ lii} \\
\end{tabular}
\caption{The Czech word \textit{`Austrálii'} is generating using three tokens with 5k BPE operations. After the 10k vocab is introduced, the model is immediately able to generate the new merged \textit{`Austrá@@'} embedding.}
\label{australii example}
\end{figure}

To quantify new vocabulary usage, we compute the F1 of newly introduced vocabulary at each increment. 
We compare the \textbf{AE} with two systems from the sweep: 20k BPE as the best sweep system, and 40k to include the full vocab used by the incremental system.
We see that after new vocab is introduced, F1 quickly improves for those new subwords over the next 3-4 epochs.
The incremental system reaches a higher F1 than the 20k system on words from the 10k and 20k vocabularies. 
Compared to the 40k system, the incremental system decreases slightly (1-1.5 F1) in performance on subwords in the 10k set, but improves on the less frequent 20k, 30k, \& 40k vocab by 3, 4, \& 5 F1, respectively.
This indicates the incremental system learns to generate introduced vocabulary comparably to the constant-vocab sweep experiments, and may even better predict less frequent vocabulary.

\subsection{Use of Embedding Inventories over Time}

BPE incrementally merges subwords to form new vocabulary items.
This means that some of the subwords from smaller BPE vocabularies may be subsumed by later merges, e.g. vocabulary items from 10k BPE operations may no longer be used with 20k or 30k.
We investigate if some embeddings become `obsolete' during our incremental BPE training, or rather, are no longer generated for \texttt{dev} and \texttt{test}, and could potentially be removed from the embedding layer. 
To see if this is the case, we compare the vocabularies generated by our \textbf{AE} model for \texttt{dev} and \texttt{test} before and after additional vocabulary is added to see if they are no longer generated after this point.
With the 10k inventory, \texttt{dev}+\texttt{test} has 5545 unique types. 
After moving to the 20k vocabulary, 26\% of these are no longer generated. 
Similarly, when incrementing from 20k to 30k, 26\% of the previous embeddings are no longer generated. 
In addition, we see that ${\sim}50$\% of the `obsolete' subwords are also not present in the equivalent references for the next increment. For example, half of the subwords that are no longer generated after the 30k vocabulary is added are also not in the 30k BPE'd reference.
For the remaining half, this does not necessarily mean that certain words can no longer be generated correctly, but that they may use a different BPE inventory to do so, as we preliminarily explore below.
We keep all embeddings added during training, but these numbers suggest some are no longer needed and could reasonably be dropped.

After incrementing, some subwords that have been subsumed by subsequent merges are nonetheless still generated by our model.
For instance, after moving from the 10k to 20k vocabulary, 7\% of the unique subwords generated by the system come from the 10k set and are not present in the 20k reference due to subsequent merges. 
Similarly, when moving from 20k to 30k, 12\% of the unique subwords generated come from the 20k set but do not appear in the 30k reference.
If training directly using a constant 20k or 30k BPE vocabulary, these subwords would not be present. However, with both options available, our network sometimes makes use of smaller subwords. 
This suggests different granularities may be optimal for different purposes, and allowing the network to combine them, as in \cite{luong2016character}, may yield further improvements. 
Future work could look into what makes different granularities optimal at different times, and extending this method to segmentation schemes beyond BPE.

\section{Related Work}

In addition to BPE, a large body of work has been dedicated to the task of segmenting words to reduce vocabulary sparsity. 
Some techniques have leveraged morphological analysis \cite{bojar2007english,subotin2011exponential,huck2017target,tamchyna2017modeling}.
Other approaches have incorporated character-based models, either as a primary model or alongside a word or subword-based model, to handle rare and unknown words \cite{ling2015finding,chung2016character,luong2016character}. 
Each of these methods introduces new complexity and challenges. 
Systems utilizing morphological inflection often require additional linguistic resources, while character-based models can be difficult to train and produce non-existent words \cite{ataman2017linguistically,neubig2012machine}. 
As a result, BPE has become the de-facto standard for segmentation and so the focus of our work.

Our methods of expanding decoder and target side embedding layers are similar to previous techniques for dynamically modifying network capacity \cite{chen2015net2net,wang2017brain,lee2017lifelong}.
Much of the previous work in this area has focused on adapting a previously trained network to new or significantly different tasks. 
Our work differs in two important ways. 
First, our emphasis is on optimizing the segmentation granularity of the input to improve translation performance for a particular task, as opposed to adapting the network to a new task. 
Concurrent to this work, \cite{kudo2018subword} also uses multiple segmentation granularities during training, but to introduce noise for robustness. 
Second, our system focuses on optimizing a hyperparameter through a single continuous training process, as opposed to multiple disconnected ones.

\section{Conclusion}

While BPE has become the standard subword approach for NMT, tuning the number of subword units is not standard practice due to the resource cost.
We see here however that the granularity of subword units can depend on language and dataset, and the choice of this parameter has greater impact with more training data. 
We presented a novel method to tune subword granularity during a single training pass, without an expensive parameter search, by incrementally introducing new vocabulary online based on \texttt{dev} loss.
Our method is able to match the best BLEU found with grid search, shown across two languages and dataset sizes. 
Further, by essentially `pretraining' with smaller, more frequent subwords, our online training method improves the translation accuracy of rare words and leads to earlier convergence.
In conclusion, our method reliably tunes subword granularity in a single training run, yielding the benefits of a sweep without the cost.

\bibliography{bibliography}
\bibliographystyle{aaai}
\end{document}